\documentclass{article}

\usepackage{PRIMEarxiv}

\usepackage[numbers,square,sort&compress]{natbib}

\usepackage[utf8]{inputenc} 
\usepackage[T1]{fontenc}    
\usepackage{booktabs}       
\usepackage{amsfonts}       
\usepackage{nicefrac}       
\usepackage{microtype}      
\usepackage{lipsum}         
\usepackage{fancyhdr}       
\usepackage{graphicx}       

\graphicspath{{media/}}     

\usepackage{hyperref} 

\hypersetup{
    colorlinks=true,
    linkcolor=blue,
    citecolor=blue,
    urlcolor=blue
}

\title{A Machine Learning Approach for Detection of Mental Health Conditions and Cyberbullying from Social Media
\thanks{
\textbf{Oral Presentation at the AAAI-26 Bridge Program on AI for Medicine and Healthcare (AIMedHealth)}. To appear in \textit{Proceedings of Machine Learning Research (PMLR)}.} 
}

\author{
  Edward Ajayi, Martha Kachweka, Mawuli Deku, Emily Aiken \\
  Carnegie Mellon University Africa \\
  Kigali, Rwanda \\
  \texttt{\{eaajayi, mkachwek, mdeku, eaiken\}@andrew.cmu.edu} \\
}

\begin{document}
\maketitle

\begin{abstract}
Mental health challenges and cyberbullying are increasingly prevalent in digital spaces, necessitating scalable and interpretable detection systems. This paper introduces a unified multiclass classification framework for detecting ten distinct mental health and cyberbullying categories from social media data. We curate datasets from Twitter and Reddit, implementing a rigorous 'split-then-balance' pipeline to train on balanced data while evaluating on a realistic, held-out imbalanced test set. We conduct a comprehensive evaluation comparing traditional lexical models, hybrid approaches, and several end-to-end fine-tuned transformers. Our results demonstrate that end-to-end fine-tuning is critical for performance, with the domain-adapted MentalBERT emerging as the top model, achieving an accuracy of 0.92 and a Macro F1 score of 0.76, surpassing both its generic counterpart and a zero-shot LLM baseline. Grounded in a comprehensive ethical analysis, we frame the system as a human-in-the-loop screening aid, not a diagnostic tool. To support this, we introduce a hybrid SHAP-LLM explainability framework and present a prototype dashboard ("Social Media Screener") designed to integrate model predictions and their explanations into a practical workflow for moderators. Our work provides a robust baseline, highlighting future needs for multi-label, clinically-validated datasets at the critical intersection of online safety and computational mental health.
\end{abstract}


\section{Introduction}
Mental health disorders are a growing global concern, affecting one in eight people worldwide \cite{WHO}. Common mental health conditions like anxiety, depression, bipolar disorder, and stress-related illnesses contribute substantially to the global burden of disease. Despite the availability of effective prevention and treatment options, access to mental health care remains a major challenge, exacerbated by stigma, discrimination, and inadequate resources \cite{WHO}.

Social media platforms contribute to and mediate mental health conditions in an increasingly digital world. While social media sites have been shown to increase connection in some settings \cite{zsila2023}, they have also amplified mental health risks by exposing users to cyberbullying \cite{naslund2020}, emotionally charged content \cite{acm3653303}, and negative sentiment \cite{wu2025}. Social media also provides a valuable opportunity for scalable screening and content moderation, with the possibility of automated detection of signs of distress through natural language processing (NLP) techniques \cite{mobin}. While existing research has made substantial progress in predicting mental health conditions from social media posts, most studies focus on binary classification (e.g., classifying posts as depression or not depression) \cite{kumar2022}, overlooking the interconnected nature of mental health conditions \cite{WHO}. Moreover, existing research often focuses on narrow subsets of harmful content, failing to address the diverse range of mental health expressions and cyberbullying behaviors encountered online.

This study aims to fill this gap by developing and evaluating a unified multiclass classification framework for detecting ten distinct categories of mental health conditions and cyberbullying from public social media data. We aggregate datasets from Reddit and Twitter, comparing lexical (TF-IDF) and contextual (e.g., BERT) modeling approaches. We explicitly frame this framework not as a diagnostic tool, but as a human-in-the-loop screening aid intended for trained moderators. To support this practical integration, we introduce a hybrid SHAP-LLM explainability system and present a prototype dashboard ("Social Media Screener") to visualize how our model's outputs can be safely and transparently operationalized.
 
We propose a dual-purpose application: (1) as a content flagging tool for acute-risk classes like 'Suicide' that require urgent human review, and (2) as a component in longitudinal analysis tools to help practitioners identify linguistic patterns over time for nuanced conditions like 'Bipolar Disorder'. This human-in-the-loop framework serves as a direct parallel to clinical support systems, where similar AI tools can assist therapists in reviewing high volumes of patient-generated data. Our evaluation, conducted on a held-out, imbalanced test set, demonstrates the effectiveness of end-to-end fine-tuning, with the domain-adapted MentalBERT \cite{ji2022mentalbert} showing clear superiority.
By integrating these technical findings with a concrete and ethically-grounded application, this study provides a comprehensive methodological baseline. Our analysis also highlights the critical limitations of existing public, weakly-labeled, single-label datasets, demonstrating the urgent need for future work in developing multi-label benchmarks sourced from consented, clinically-assessed cohorts for computational mental health.

\section{Related Work}

\subsection{Mental Health Detection from Social Media}
A large and growing body of research leverages natural language processing (NLP) to identify mental health conditions from social media platforms such as Twitter and Reddit \cite{mobin,acm3653303, ieee10803545, abdullah2024, perez2024}. Early studies often focused on binary classification tasks distinguishing between depressed and non-depressed users \cite{kumar2022}. More recent papers have incorporated deep learning to capture nuanced emotional patterns \cite{jiang2020deeppsy}. Transformer-based architectures, particularly BERT \cite{devlin2018bert}, have become the de facto standard for text classification due to their capacity to capture contextual dependencies and semantics at a fine-grained level. \cite{ji2022mentalbert} introduced MentalBERT and MentalRoBERTa, domain-adapted versions of BERT fine-tuned on Reddit mental health forums, showing notable improvements in classifying user mental health status. However, most of these studies rely on label sets with limited numbers of classes, failing to capture the complex and interrelated nature of mental health conditions in the real world. 

\subsection{Cyberbullying Detection}
Cyberbullying detection on social media has also received considerable attention from researchers in the recent years \cite{Mathew2020, antypas2023}. Most relevantly to our work, \cite{sosnet2020} developed SOSNet, a domain-specific neural network architecture for distinguishing types of cyberbullying such as those based on gender, ethnicity, and religion. We build on the dataset curated by \cite{sosnet2020} in this paper.

\section{Data and methods}

\subsection{Data}

This study aggregates a number of datasets from Twitter and Reddit to analyze and classify various mental health conditions, including suicidal ideation, personality disorders, stress, anxiety, and bipolar disorder. We also work with data on cyberbullying data from Twitter. All datasets were sourced from Kaggle.

\subsubsection{Mental health data}
The mental health data are sourced from various social media platforms, primarily Twitter and specific subreddits dedicated to mental health conversations \cite{mentaldata, redditmental}. The dataset is categorized into four conditions: anxiety, stress, bipolar disorder, and personality disorder, totaling 53,043 posts. The anxiety data was gathered from Facebook and Twitter and subsequently manually annotated by four undergraduate English-speaking students. The data for bipolar disorder and personality disorder were collected from their respective subreddits. Similarly, the stress data was sourced from relevant subreddits; however, the specific annotation guidelines for this subset were not detailed by the original authors.

\subsubsection{Cyberbullying data}
The cyberbullying dataset was adopted from \cite{sosnet2020}, who developed a model to detect cyberbullying based on age, gender, ethnicity, and religion on Twitter. The dataset labels each post as one of five types of cyberbullying (gender, religion, ethnicity, age, other) or as not containing cyberbullying content. The fine-grained labels were initially created through a manual annotation process on a subset of the data. To expand the dataset, the authors \cite{sosnet2020} then employed a semi-supervised method, Dynamic Query Expansion (DQE), to increase the number of samples for each class. The final curated dataset contains a total of 47,692 posts.

\subsubsection{Suicide and depression detection data}
Our third and final dataset focuses on detecting depression and suicidal ideation, with data collected from the SuicideWatch and depression subreddits \cite{suicidalkaggle}. The dataset also includes posts from the teenagers subreddit for non-suicidal and non-depression content. The dataset includes 232,074 posts in total, each labeled as suicidal or non-suicidal content. 

\subsubsection{Data cleaning}
All text entries were standardized to lowercase strings. Unwanted elements such as URLs, user mentions, and non-alphanumeric characters were removed using regular expressions, and extraneous whitespace was stripped to ensure data uniformity.

\subsubsection{Dataset Curation, Splitting, and Balancing}
To ensure a robust evaluation and prevent data leakage, we implemented a strict "split-then-balance" pipeline. First, all ten datasets were merged into a master dataset of 274,150 posts. We then performed post-level deduplication to prevent identical posts from appearing in both training and test sets. After deduplication, the dataset was split into a training pool (80\%) and a held-out test pool (20\%) using a stratified split to preserve the original, imbalanced class distribution in both pools. The final test set was sampled from the held-out test pool and retained its original, highly imbalanced distribution to reflect real-world data.

The final training set was constructed exclusively from the 80\% training pool using a multi-step balancing process. This process, summarized in Table \ref{tab:combined_distribution}, involved downsampling (DS) high-resource classes (e.g., Non-Suicide, Suicide) and applying deduplication (DD) and EDA-based oversampling (a technique that generates synthetic text by applying one of four random operations: synonym replacement, random insertion, random swap, or random deletion)\cite{wei2019eda} to low-resource classes (e.g., Personality Disorder, Stress). Table \ref{tab:combined_distribution} shows the class distribution after each step and the final test set.

\subsection{Data Label Verification}
\label{subsec:label_verification}

To evaluate the reliability of the dataset’s weak labels, we performed a manual annotation study on a randomly selected subset of 300 posts (1\% of the dataset), balanced across all ten classes. Two authors independently annotated the posts while blinded to the original labels. Inter-annotator agreement achieved a Cohen’s Kappa of \textbf{$\kappa = 0.76$}, indicating substantial consistency between annotators. We further compared each annotator’s labels with the original dataset labels, yielding Cohen’s Kappa scores of \textbf{$\kappa = 0.71$} and \textbf{$\kappa = 0.94$}, respectively. These results demonstrate that the weak labels are highly aligned with human judgment, supporting the reliability of the dataset for our experiments.

\begin{table}[htbp]
\centering
\setlength{\tabcolsep}{8pt} 
\renewcommand{\arraystretch}{1.0} 
\caption{Class distribution across training preprocessing steps and final test set. DS = Downsampling, DD = Deduplication, CB = Cyberbullying.}
\label{tab:combined_distribution}
\begin{tabular}{lrrrr} 
\toprule
\textbf{Class} & \textbf{Before DS} & \textbf{After DS} & \textbf{After EDA \& DD} & \textbf{Final Test Set} \\
\midrule
Age CB & 7,992 & 2,400 & 2,400 & 175 \\
Anxiety & 3,841 & 2,400 & 2,400 & 80 \\
Bipolar & 2,777 & 2,001 & 2,400 & 55 \\
Ethnicity CB & 7,955 & 2,400 & 2,400 & 172 \\
Gender CB & 7,916 & 2,400 & 2,400 & 167 \\
Non-Suicide & 115,983 & 2,400 & 2,400 & 2,549 \\
Personality Disorder & 1,077 & 714 & 2,387 & 20 \\
Religion CB & 7,997 & 2,400 & 2,400 & 175 \\
Stress & 2,585 & 1,832 & 2,396 & 50 \\
Suicide & 116,027 & 2,400 & 2,400 & 2,557 \\
\bottomrule
\end{tabular}
\end{table}

\subsection{Featurization}

We experiment with two methods to extract numerical features from the preprocessed text:

\begin{itemize}
    \item \textbf{TF-IDF Vectorization}: The Term Frequency-Inverse Document Frequency (TF-IDF) approach \cite{ieee10803545, tfidf} was implemented using the \texttt{TfidfVectorizer} from scikit-learn, configured to extract the top \textbf{5000} features. This transformation generated a sparse matrix of TF-IDF scores, emphasizing words that are important relative to their document frequency across datasets. We clarify that stopwords were removed only for exploratory lexical analysis, for all model training (both TF-IDF and BERT embeddings-based), stopwords were kept to preserve full semantic context and ensure a fair comparison.
    
    \item \textbf{BERT Embeddings}: For deeper semantic representation, BERT embeddings \cite{arxiv2403.10750} were obtained using the \texttt{bert-base-uncased} model from the Hugging Face Transformers library. Each text sample was tokenized with a maximum length of 128 tokens. 
\end{itemize}

These feature extraction strategies allowed the study to experiment with both lexical (TF-IDF) and contextual (BERT) representations.

\subsection{Machine learning approaches}
\label{sec:model_training}
\label{subsec:model_architecture}

We test several distinct families of machine learning models for our 10-class classification task. All models were trained on the balanced training set (Table \ref{tab:combined_distribution}) and evaluated on the held-out, imbalanced test set. Specific hyperparameters (Appendix \ref{sec:appendix_hyperparams}) were selected based on standard practices.

\subsubsection{Lexical Baseline Models}
To establish a non-contextual baseline, we trained two classical models on TF-IDF features. As described in Section 3.2, we used the top 5000 features (unigrams and bigrams), with stopwords preserved.

\begin{itemize}
    \item \textbf{TF-IDF + Logistic Regression:} We used scikit-learn’s \texttt{LogisticRegressionCV} with the \texttt{saga} solver, a multinomial setting, and \texttt{cv=5} (5-fold cross-validation) on the training set to select the best regularization strength.
    
    \item \textbf{TF-IDF + Support Vector Machine (SVM):} We used scikit-learn’s \texttt{GridSearchCV} with \texttt{cv=5} to find the optimal regularization parameter for a linear \texttt{SVC} class, maximizing for macro F1 score.
\end{itemize}

\subsubsection{Static Embedding Baseline Models}
To test the performance of static contextual embeddings, we fed pre-computed BERT embeddings into classical and neural classifiers. For these models, sentence-level embeddings were extracted using the CLS token from the \texttt{bert-base-uncased} model.

\begin{itemize}
    \item \textbf{BERT Embeddings + Logistic Regression:} The static CLS embeddings were fed into the same \texttt{LogisticRegressionCV} model used in the lexical baseline to ensure a fair comparison.

    \item \textbf{BERT Embeddings + RNN:} A simple Recurrent Neural Network (RNN) with one hidden layer of 128 units was built using PyTorch. The model was trained for five epochs using the Adam \cite{kingma2017adammethodstochasticoptimization} optimizer (learning rate = 0.001) on the static embeddings.

    \item \textbf{BERT Embeddings + DNN:} A feed-forward Deep Neural Network (DNN) was constructed with two hidden layers (256 and 128 units) with ReLU activations and dropout. This model was also trained for five epochs with the Adam optimizer (learning rate = 0.001) on the static embeddings.
\end{itemize}

\subsubsection{End-to-End Fine-Tuned Transformer Models}
This group represents our primary end-to-end models, where the entire transformer architecture is updated during training. For these, we used a 90/10 split on our main training set for training and validation, respectively. All models were fine-tuned using the AdamW optimizer \cite{loshchilov2019decoupledweightdecayregularization} with a learning rate of 2e-5 and a batch size of 16. We compared four different architectures:

\begin{itemize}
    \item \textbf{Finetuned BERT-base:} The \texttt{bert-base-uncased} model \cite{devlin2019bert}, fine-tuned end-to-end. This serves as our primary contextual model.
    
    \item \textbf{Finetuned MentalBERT:} A domain-adapted model \cite{ji2022mentalbert} pre-trained on text from Reddit mental health forums.
    
    \item \textbf{Finetuned MentalRoBERTa:} A RoBERTa-base model also pre-trained on the same domain-specific mental health corpus as MentalBERT\cite{ji2022mentalbert}.
    
    \item \textbf{Finetuned ModernBERT:} We benchmark ModernBERT~\cite{warner2024smarterbetterfasterlonger} to test whether recent architectural improvements offer performance gains over standard BERT on our task, even without domain-specific pre-training.
\end{itemize}

\subsubsection{Zero-Shot LLM Baseline}
As a modern, non-finetuned baseline, we evaluated a powerful Large Language Model (GPT-OSS 120B)\cite{agarwal2025gpt} in a zero-shot setting\cite{kojima2022large}. We prompted the model to classify samples from our test set into one of the ten categories. This approach required a post-processing step to map the model's textual outputs to our valid class labels; approximately 31.6\% of responses did not map to a valid label and were excluded from the LLM's performance calculation.

\subsection{Evaluation Metrics}
\label{subsec:eval_metrics}

All performance metrics are reported on the held-out, imbalanced test set (Table \ref{tab:combined_distribution}). This ensures our evaluation realistically assesses model performance on the original, real-world class distribution.

We report the following metrics of model performance:
\begin{itemize}
    \item \textbf{Accuracy}: Overall proportion of correct predictions.
    
    \item \textbf{Macro F1}: The unweighted average of all F1 scores across classes. The macro F1 score treats all classes equally, which is critical for highlighting performance on our rare, low-resource classes. 
    
    \item \textbf{Weighted F1}: The average of the per-class F1 scores, weighted by the size (support) of the class in the dataset. 
\end{itemize}

We additionally assess the performance of each ML model in each class, to identify which classes are more and less challenging to predict. We report the following per-class metrics of model performance:
\begin{itemize}
    \item \textbf{Precision \& Recall}: Precision is the proportion of correct positive predictions; recall is the proportion of actual positives identified as positive.
    
    \item \textbf{Per-Class F1}: F1 score reported individually per class. The F1 score is the harmonic mean of precision and recall.
    
    \item \textbf{AUPRC:} For the high-risk \textit{Suicide} class, we additionally report the Area Under the Precision–Recall Curve~\cite{mcdermott2024closer}, which is more informative under class imbalance.
    
    \item \textbf{Calibration Plot}: We also provide a calibration analysis for the 'Suicide' class to assess whether the models' predicted confidence scores are reliable.
\end{itemize}

\section{Exploratory data analysis}

\begin{figure}[h]
    \centering
    \includegraphics[width=0.99\linewidth]{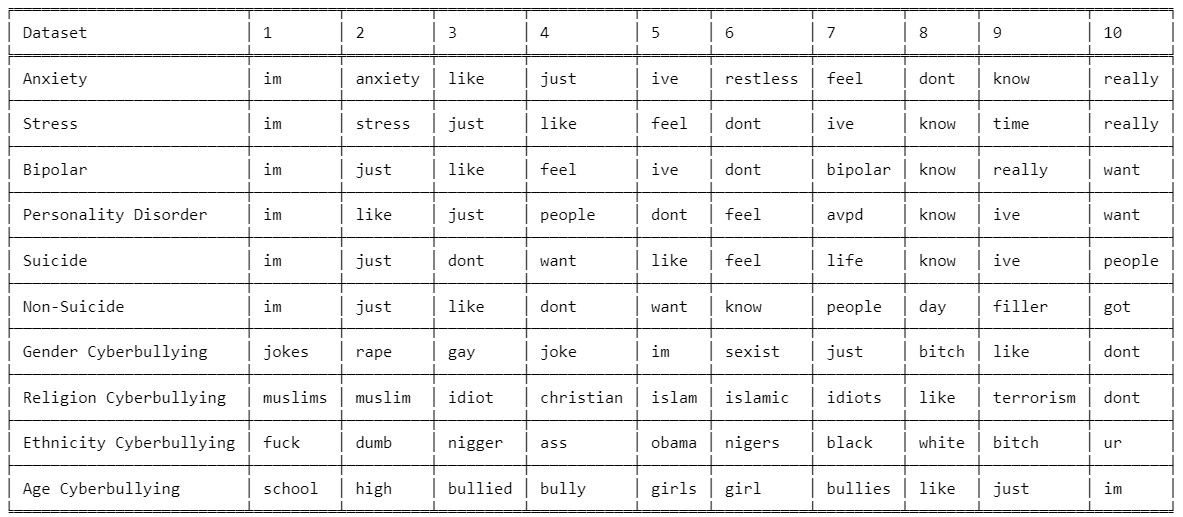}
    \caption{Table showing top 10 TF-IDF words for each class label}
    \label{fig:idf_words}
\end{figure}

We conducted both raw word frequency analysis and TF-IDF (Term Frequency–Inverse Document Frequency) analysis to identify representative words within each dataset. While the raw frequency approach consistently surfaced high-frequency function words such as “I”, “the”, “to”, and “my”,  TF-IDF assigns higher importance to words that are characteristic of a specific category but rare in others. This approach enabled us to uncover more semantically meaningful and category-specific terms such as bipolar (bipolar class), rape (gender cyberbullying class), and bullied (age cyberbullying class). Figure \ref{fig:idf_words} presents the top ten most frequent words associated with each class label in the dataset. 

\begin{figure}[htbp]
    \centering
    \includegraphics[width=0.9\linewidth]{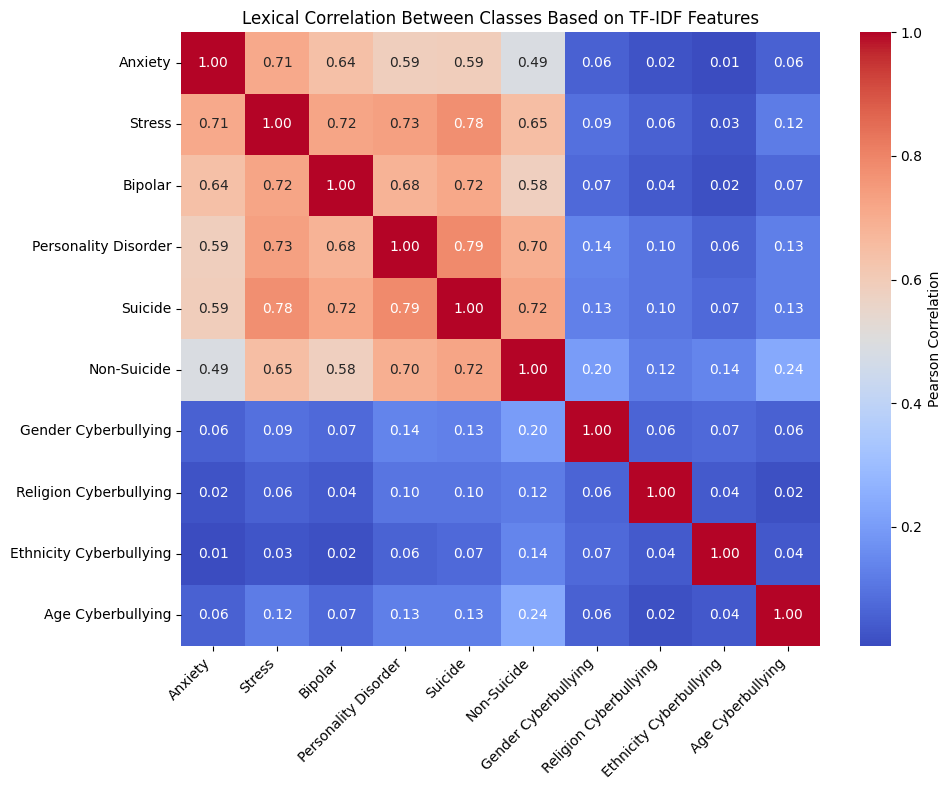 }
    \caption{Plot of correlation of TF-IDF embeddings across different class labels}
    \label{fig:corr_tfidf}
\end{figure}

To examine the lexical similarity between classes, we performed a correlation analysis of the TF-IDF features. Inspired by past natural language processing work taking similar approaches to compare classes \cite{mobin, liu_sentence_2020}, we computed the mean TF-IDF vector by averaging across all its documents. The Pearson correlation coefficient was then computed between the mean TF-IDF vectors of each dataset pair to assess the degree of lexical similarity. The resulting heatmap is presented in Figure~\ref{fig:corr_tfidf}. High positive correlations indicate shared vocabulary patterns, while low correlations suggest distinct linguistic structures. Strong correlations are observed among mental health-related classes such as stress, bipolar, and personality Disorder. In contrast, cyberbullying classes exhibit much lower correlations with mental health datasets, indicating distinct linguistic patterns. 

\section{Results}
\label{sec:results}

This section presents the empirical results of our experiments. We first provide a comparative analysis of the overall performance of all models, followed by a detailed per-class breakdown to identify specific strengths and weaknesses. Finally, we conduct a focused analysis on the critical task of suicide detection, evaluating model reliability through AUPRC and calibration plots.

\subsection{Overall Model Performance}

Our results demonstrate a clear performance advantage for end-to-end fine-tuned transformer models over both traditional machine learning (ML) methods and hybrid approaches that use static BERT embeddings. As shown in Table \ref{tab:model_performance}, \textbf{Finetuned MentalBERT} emerged as the top-performing model, achieving an accuracy of 0.92 and a Macro F1 score of 0.76. The other fine-tuned variants, including the generic BERT-base, RoBERTa, and ModernBERT, also delivered strong performance, with Macro F1 scores ranging from 0.70 to 0.71. The competitive performance of the generic fine-tuned BERT highlights the effectiveness of our data curation and balancing pipeline in adapting a general-purpose model to this specific task.

In contrast, the traditional and hybrid ML models exhibited lower performance (Table \ref{tab:ml_model_performance}). The best-performing ML model, TF-IDF\_LogReg, achieved a Macro F1 score of 0.67. Models using static BERT embeddings as features for classical classifiers (BERT\_LogReg, BERT\_RNN, BERT\_DNN) performed the poorest, with Macro F1 scores between 0.53 and 0.58. This significant gap underscores the necessity of fine-tuning the entire transformer architecture to capture the complex contextual nuances present in the data.

\begin{table}[h]
\centering
\setlength{\tabcolsep}{4pt} 
\caption{Overall Model Performance Comparison for Finetuned BERT variants.}
\label{tab:model_performance}
\begin{tabular}{lccccc}
\hline
\textbf{Model} & \textbf{Accuracy} & \textbf{Macro F1} & \textbf{Weighted F1} & \textbf{Precision} & \textbf{Recall} \\
\hline
BERT & 0.87 & 0.70 & 0.88 & 0.64 & 0.88 \\
MentalBERT & 0.92 & 0.76 & 0.93 & 0.70 & 0.89 \\
RoBERTa & 0.88 & 0.70 & 0.90 & 0.64 & 0.90 \\
ModernBERT & 0.89 & 0.71 & 0.91 & 0.64 & 0.88 \\
\hline
\end{tabular}
\end{table}

\begin{table}[h!]
\centering
\setlength{\tabcolsep}{4pt} 
\caption{Overall ML Model Performance Comparison.}
\label{tab:ml_model_performance}
\begin{tabular}{lccccc}
\hline
\textbf{Model} & \textbf{Accuracy} & \textbf{Macro F1} & \textbf{Weighted F1} & \textbf{Precision} & \textbf{Recall} \\
\hline
TF-IDF\_LogReg & 0.82 & 0.67 & 0.84 & 0.61 & 0.82 \\
TF-IDF\_SVM & 0.80 & 0.64 & 0.83 & 0.58 & 0.81 \\
BERT\_LogReg & 0.75 & 0.55 & 0.79 & 0.49 & 0.76 \\
BERT\_RNN & 0.68 & 0.53 & 0.74 & 0.48 & 0.74 \\
BERT\_DNN & 0.77 & 0.58 & 0.80 & 0.51 & 0.76 \\
\hline
\end{tabular}
\end{table}

\subsection{Per-Class Performance Analysis}

A granular, per-class analysis reveals significant performance disparities across the ten categories, reflecting the challenges posed by our realistic, imbalanced test set (Table \ref{tab:finetuned_metrics}). The fine-tuned models, particularly MentalBERT, excelled on high-signal categories with explicit lexical cues. For instance, MentalBERT achieved outstanding F1-scores on cyberbullying classes like \textit{Age CB} (0.95) and \textit{Religion CB} (0.96), as well as the critical mental health class \textit{Suicide} (0.96).

Conversely, performance was substantially lower for nuanced mental health conditions that were sparsely represented in the imbalanced test set and are characterized by greater lexical ambiguity. The F1-scores for \textit{Personality Disorder} (0.32), \textit{Stress} (0.46), and \textit{Bipolar} (0.70) were markedly lower for MentalBERT. This outcome is not an indication of model failure but rather a realistic reflection of the inherent difficulty of detecting these conditions from isolated posts without longitudinal context. The severe class imbalance in the test set, which mirrors real-world data distribution, correctly penalizes models for misclassifying these rare but important cases, leading to lower but more credible Macro F1 scores. The traditional ML models followed a similar trend but with uniformly lower scores across all classes (Table \ref{tab:ml_metrics}).

\begin{table}[h]
\centering
\scriptsize
\caption{Per-Class F1, Precision, and Recall – Fine-tuned Models.}
\label{tab:finetuned_metrics}
\begin{tabular}{lcccccccccc}
\hline
\textbf{Model} & \textbf{Age CB} & \textbf{Anx.} & \textbf{Bip.} & \textbf{Eth. CB} & \textbf{Gen. CB} & \textbf{Non-Su.} & \textbf{Pers. Dis.} & \textbf{Rel. CB} & \textbf{Str.} & \textbf{Suic.} \\
\hline
\multicolumn{11}{l}{\textbf{F1-Score}} \\
\hline
BERT       & 0.92 & 0.71 & 0.62 & 0.87 & 0.64 & 0.85 & 0.19 & 0.93 & 0.30 & 0.96 \\
MentalBERT & 0.95 & 0.65 & 0.70 & 0.91 & 0.74 & 0.93 & 0.32 & 0.96 & 0.46 & 0.96 \\
RoBERTa    & 0.92 & 0.60 & 0.40 & 0.89 & 0.68 & 0.88 & 0.17 & 0.94 & 0.56 & 0.96 \\
ModernBERT & 0.91 & 0.66 & 0.61 & 0.87 & 0.71 & 0.90 & 0.24 & 0.94 & 0.33 & 0.96 \\
\hline
\multicolumn{11}{l}{\textbf{Precision}} \\
\hline
BERT       & 0.88 & 0.61 & 0.50 & 0.79 & 0.48 & 0.98 & 0.11 & 0.89 & 0.18 & 0.96 \\
MentalBERT & 0.96 & 0.52 & 0.62 & 0.86 & 0.61 & 0.97 & 0.21 & 0.95 & 0.32 & 0.99 \\
RoBERTa    & 0.88 & 0.45 & 0.26 & 0.83 & 0.52 & 0.98 & 0.10 & 0.91 & 0.46 & 0.98 \\
ModernBERT & 0.86 & 0.52 & 0.48 & 0.78 & 0.57 & 0.97 & 0.14 & 0.91 & 0.22 & 0.99 \\
\hline
\multicolumn{11}{l}{\textbf{Recall}} \\
\hline
BERT       & 0.97 & 0.86 & 0.82 & 0.97 & 0.95 & 0.75 & 0.75 & 0.98 & 0.82 & 0.96 \\
MentalBERT & 0.95 & 0.89 & 0.80 & 0.97 & 0.93 & 0.89 & 0.70 & 0.97 & 0.82 & 0.94 \\
RoBERTa    & 0.97 & 0.91 & 0.87 & 0.95 & 0.97 & 0.79 & 0.90 & 0.98 & 0.74 & 0.94 \\
ModernBERT & 0.97 & 0.90 & 0.84 & 0.98 & 0.95 & 0.84 & 0.65 & 0.98 & 0.74 & 0.93 \\
\hline
\end{tabular}%
\\[4pt]
\textit{Note:} CB—Cyberbullying, Anx.—Anxiety, Bip.—Bipolar, Eth.—Ethnicity, Gen.—Gender, Non-Su.—Non-Suicidal, Pers. Dis.—Personality Disorder, Rel.—Religion, Str.—Stress, Suic.—Suicide.
\end{table}

\begin{table}[htbp]
\centering
\scriptsize
\caption{Per-class F1, Precision, and Recall for ML Models.}
\label{tab:ml_metrics}
\begin{tabular}{lcccccccccc}
\hline
\textbf{Model} & \textbf{Age CB} & \textbf{Anx.} & \textbf{Bip.} & 
\textbf{Eth. CB} & \textbf{Gen. CB} & \textbf{Non-Su.} &
\textbf{Pers. Dis.} & \textbf{Rel. CB} & \textbf{Str.} & \textbf{Suic.} \\
\hline
\multicolumn{11}{l}{\textbf{F1-Score}} \\
\hline
TF-IDF\_LogReg & 0.89 & 0.57 & 0.44 & 0.86 & 0.67 & 0.85 & 0.43 & 0.93 & 0.19 & 0.87 \\
TF-IDF\_SVM    & 0.92 & 0.56 & 0.39 & 0.86 & 0.62 & 0.83 & 0.28 & 0.93 & 0.17 & 0.85 \\
BERT\_LogReg   & 0.73 & 0.43 & 0.26 & 0.72 & 0.52 & 0.78 & 0.14 & 0.89 & 0.17 & 0.85 \\
BERT\_RNN      & 0.70 & 0.45 & 0.30 & 0.68 & 0.41 & 0.73 & 0.20 & 0.88 & 0.09 & 0.81 \\
BERT\_DNN      & 0.73 & 0.52 & 0.30 & 0.71 & 0.62 & 0.79 & 0.25 & 0.81 & 0.16 & 0.86 \\
\hline
\multicolumn{11}{l}{\textbf{Precision}} \\
\hline
TF-IDF\_LogReg & 0.84 & 0.43 & 0.32 & 0.80 & 0.54 & 0.89 & 0.31 & 0.89 & 0.11 & 0.93 \\
TF-IDF\_SVM    & 0.88 & 0.41 & 0.27 & 0.78 & 0.48 & 0.88 & 0.18 & 0.91 & 0.10 & 0.93 \\
BERT\_LogReg   & 0.62 & 0.30 & 0.17 & 0.60 & 0.37 & 0.91 & 0.08 & 0.83 & 0.10 & 0.93 \\
BERT\_RNN      & 0.59 & 0.31 & 0.21 & 0.55 & 0.27 & 0.93 & 0.13 & 0.83 & 0.05 & 0.95 \\
BERT\_DNN      & 0.61 & 0.40 & 0.20 & 0.58 & 0.49 & 0.91 & 0.17 & 0.69 & 0.09 & 0.93 \\
\hline
\multicolumn{11}{l}{\textbf{Recall}} \\
\hline
TF-IDF\_LogReg & 0.95 & 0.88 & 0.71 & 0.94 & 0.87 & 0.80 & 0.70 & 0.97 & 0.58 & 0.82 \\
TF-IDF\_SVM    & 0.96 & 0.89 & 0.71 & 0.97 & 0.89 & 0.79 & 0.65 & 0.96 & 0.52 & 0.79 \\
BERT\_LogReg   & 0.89 & 0.76 & 0.65 & 0.89 & 0.87 & 0.68 & 0.50 & 0.96 & 0.60 & 0.79 \\
BERT\_RNN      & 0.87 & 0.81 & 0.56 & 0.90 & 0.92 & 0.60 & 0.40 & 0.95 & 0.72 & 0.71 \\
BERT\_DNN      & 0.91 & 0.74 & 0.58 & 0.91 & 0.83 & 0.71 & 0.50 & 0.97 & 0.64 & 0.81 \\
\hline
\end{tabular}
\\[4pt]
\textit{Note:} CB—Cyberbullying, Anx.—Anxiety, Bip.—Bipolar, Eth.—Ethnicity,  
Gen.—Gender, Non-Su.—Non-Suicidal, Pers. Dis.—Personality Disorder,  
Rel.—Religion, Str.—Stress, Suic.—Suicide.
\end{table}

\subsection{Analysis on Suicide Detection: AUPRC and Calibration}
\begin{figure}[htbp]
    \centering
    \includegraphics[width=0.9\columnwidth]{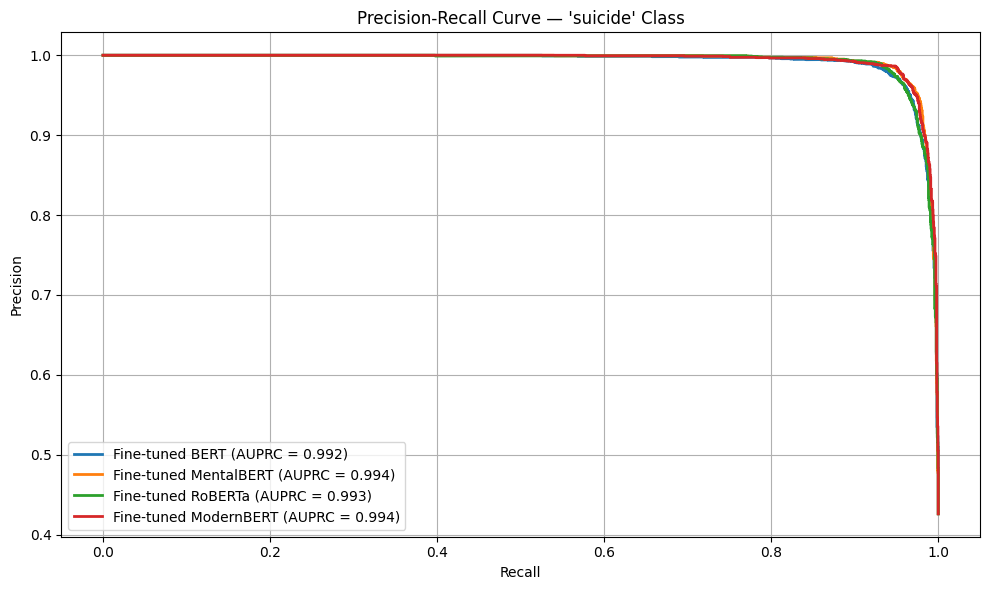}
    \caption{Precision-Recall Curve for the 'Suicide' class, showing near-perfect AUPRC scores for all fine-tuned models.}
    \label{fig:auprc_suicide}
\end{figure}

For the high-stakes task of suicide detection, we evaluated model reliability. All fine-tuned models demonstrated excellent discriminative power, achieving near-perfect AUPRC scores ($\ge 0.992$), with MentalBERT and ModernBERT reaching 0.994 (Figure \ref{fig:auprc_suicide}). This confirms their ability to identify suicidal content with high precision.

\begin{figure}[htbp]
    \centering
    \includegraphics[width=0.9\columnwidth]{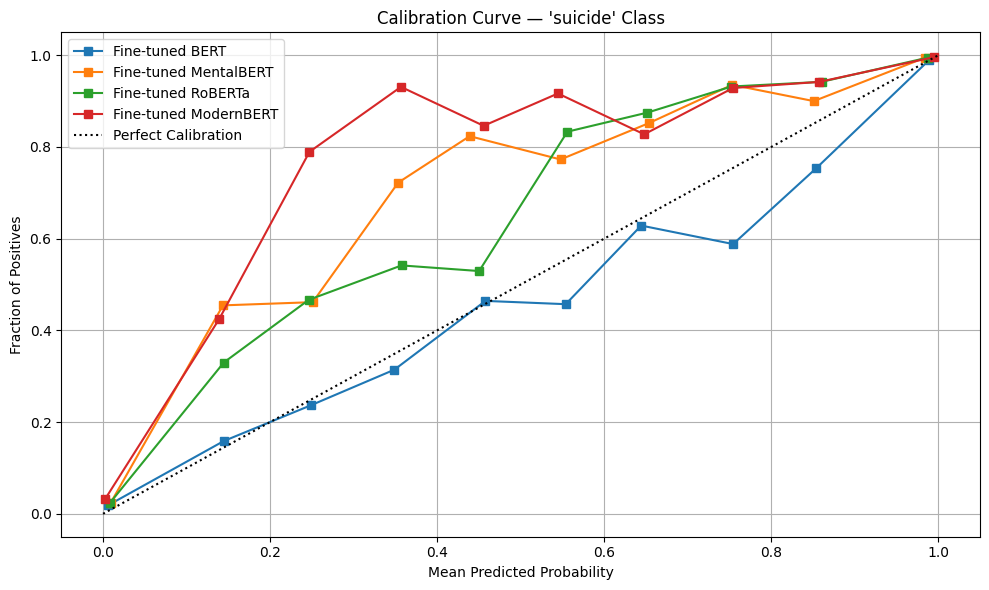}
    \caption{Calibration Curve for the 'Suicide' class. MentalBERT demonstrates the best calibration, with its predicted probabilities closely matching the observed frequencies.}
    \label{fig:calibration_suicide}
\end{figure}

Beyond discrimination, the calibration analysis (Figure \ref{fig:calibration_suicide}) identifies MentalBERT as the most trustworthy model. Its confidence scores are well-calibrated, in contrast to the under-confident standard BERT and over-confident ModernBERT. This superior reliability makes MentalBERT the most suitable model for this critical screening task.

\subsection{Impact of Data Balancing Strategy}
\label{subsec:balancing_impact}

To evaluate the effectiveness of our dual data balancing approach, we compared model performance before and after its application. The results confirm a crucial improvement in overall robustness. For our top-performing model, MentalBERT, the Macro F1 score, the most significant metric for imbalanced classes, increased from \textbf{0.73 to 0.76}, while accuracy rose from 0.90 to 0.92. This demonstrates the value of our pipeline in creating a more equitable and effective classifier.

The strategy's primary benefits are most evident at the per-class level, validating its targeted impact. The most substantial gain occurred in the \textbf{\textit{Bipolar}} class, where the F1-score rose dramatically from \textbf{0.49 to 0.70}. Notable improvements were also observed for other underrepresented mental health classes like \textbf{\textit{Anxiety}} (0.59 to 0.65) and \textbf{\textit{Personality Disorder}} (0.28 to 0.32). Interestingly, the strategy did not improve all rare classes equally; the F1-score for \textit{Stress} remained unchanged, suggesting its lexical ambiguity presents a challenge that oversampling alone cannot solve. Overall, these results justify our methodological choice, confirming that the balancing strategy provides a targeted performance lift for key underrepresented classes.

\subsection{Error Analysis}
\label{subsec:error_analysis}

\begin{figure}[htbp]
    \centering
    \includegraphics[width=0.95\columnwidth]{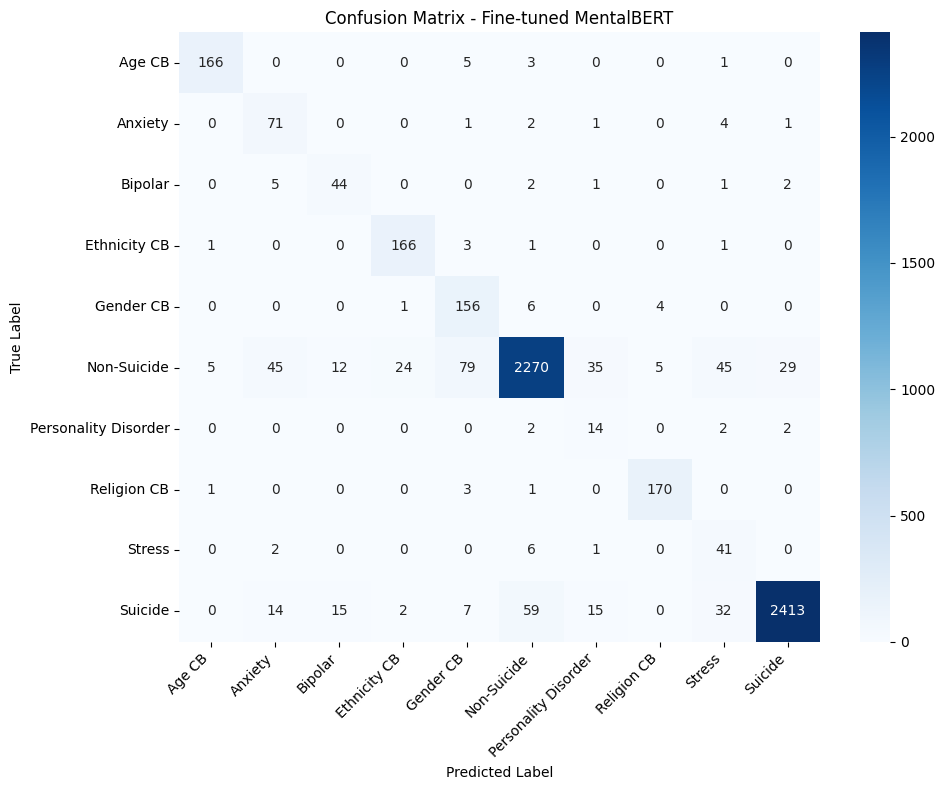}
    \caption{Confusion matrix for Fine-tuned MentalBERT. The model's primary errors occur between semantically related classes (e.g., Anxiety and Stress), confirming that it learns content over platform-specific artifacts.}
    \label{fig:confusion_matrix}
\end{figure}

An analysis of the confusion matrix for our top model, MentalBERT (Figure \ref{fig:confusion_matrix}), reveals key learning behaviors. The model demonstrates high performance on well-represented classes with clear lexical signals, such as \textit{Suicide} and the cyberbullying categories, as indicated by the strong diagonal.

More importantly, the off-diagonal error patterns provide strong evidence that the model is learning true semantic relationships, not superficial domain cues. Misclassifications are concentrated between semantically coherent categories, such as the confusion between \textit{Anxiety} and \textit{Stress}. In contrast, cross-domain errors between the mental health (primarily Reddit-based) and cyberbullying (primarily Twitter-based) categories are minimal: only 2.3\% (123 out of 5,311) of mental health posts were misclassified as a cyberbullying class, while a mere 1.5\% (10 out of 689) of cyberbullying posts were misclassified as mental health. This pattern refutes the hypothesis of the model learning spurious platform heuristics and confirms it is addressing the intended nuanced classification task.

\section{Ethical Considerations}
\label{sec:ethics}

The development of AI for mental health screening requires a rigorous ethical framework to ensure responsible innovation. Our work is therefore grounded in a framework designed to proactively address the critical challenges of model application, data provenance, and fairness.

\subsection{Intended Application and Limitations}
A primary ethical risk is the misinterpretation of model outputs as clinical diagnoses. We explicitly state that our model is not a diagnostic tool. Instead, we propose its use as a screening assistant for trained human moderators within a dual-purpose framework:
\begin{itemize}
    \item \textbf{For Immediate Risk Flagging:} For high-signal classes like \textit{Suicide}, the model can effectively flag content for urgent human review, helping to prioritize potentially life-saving interventions.

    \item \textbf{Nuanced Pattern Analysis:} For conditions like \textit{Bipolar Disorder} that require longitudinal assessment, the model supports a broader decision-support tool by highlighting linguistic shifts over time. For example, tracking changes in language associated with mood episodes can offer an early signal for clinical intervention, rather than labeling individual posts in isolation.

\end{itemize}
In all cases, the system is designed to inform, not replace, professional human judgment.

\subsection{Data Privacy and Provenance}
Our work uses anonymized, publicly available data from Kaggle. All datasets were curated in accordance with platform terms and community research standards. The data were further processed to remove any personally identifiable information, ensuring compliance with ethical use guidelines. Our label verification study (Section \ref{subsec:label_verification}) served as an additional quality assurance step to validate labeling consistency and dataset integrity.

\subsection{Accountability and Misuse}
This framework is designed for a strictly supportive purpose, operationalized through a human-in-the-loop system where trained professionals verify all automated flags. The intended application of the resulting tool is to assist content moderation and connect individuals with resources. Consequently, any use for surveillance, censorship, or punitive action would constitute a misuse and falls outside the tool's ethical scope.

\subsection{Model Explainability}
To ensure our framework is transparent and trustworthy for real-world applications, we developed a hybrid explainability system. This approach combines results from SHAP (SHapley Additive exPlanations)\cite{lundberg2017unifiedapproachinterpretingmodel} with the narrative clarity of a Large Language Model (GPT-OSS 20B)\cite{agarwal2025gpt}. First, SHAP is used to identify the precise mathematical contribution of each word to the model's prediction. This quantitative evidence is then synthesized by the LLM to generate a coherent, human-readable explanation.

\begin{figure}[h]
    \centering
    \includegraphics[width=\linewidth]{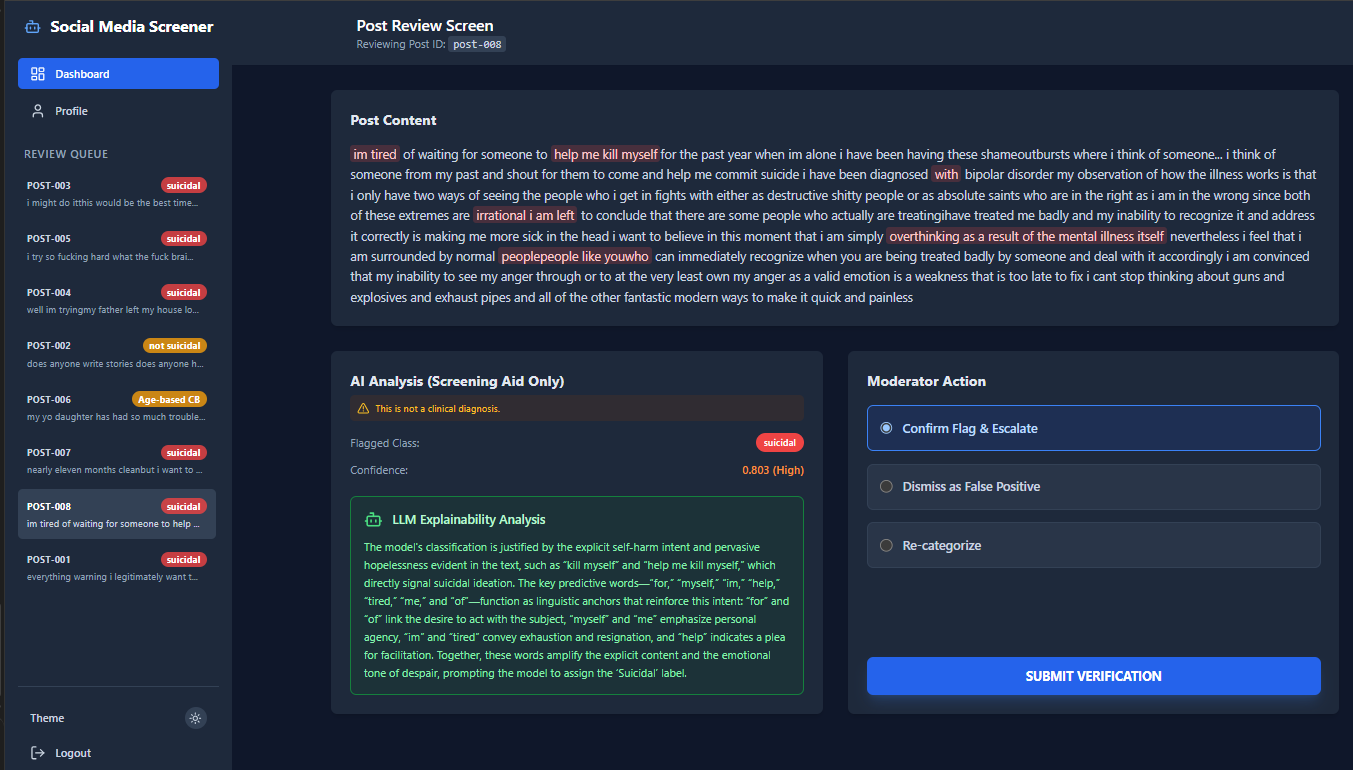} 
    \caption{The proposed "Social Media Screener" prototype. This interface integrates the hybrid explainability system directly for a human moderator. The SHAP-derived token importance is shown via red highlights in the 'Post Content' section, while the 'LLM Explainability Analysis' is presented in a clear text box below the model's flag.}
    \label{fig:shap_explain_suicide}
\end{figure}

To demonstrate how this system is integrated into our proposed human-in-the-loop application, Figure \ref{fig:shap_explain_suicide} shows a prototype of the "Social Media Screener" dashboard.

This prototype visualizes how a human moderator interacts with the system. Instead of a raw plot, the SHAP analysis is rendered as simple \textbf{red highlights} on the most impactful words (e.g., "help me kill," "myself," "tired of"). This quantitative data is paired with the LLM Explainability Analysis in the green box, which provides a narrative summary. This interface also includes the critical ethical safeguard ("This is not a clinical diagnosis.") and the "Moderator Action" panel, which requires a human to verify every flag, completing the human-in-the-loop framework. Additional examples can be found in Appendix~\ref{sec:appendix_explain}.

\subsection{Analysis of Performance Disparities and Domain Cues}
\label{subsec:performance_analysis}

Our per-class results show performance variations across categories, which our analysis attributes to the inherent characteristics of the data rather than model artifacts. As confirmed by our error analysis (Section \ref{subsec:error_analysis}), the model does not rely on superficial platform cues (e.g., Reddit vs. Twitter) but instead learns from the content's semantic properties. Performance differences are primarily driven by two factors:

\begin{itemize}
    \item \textbf{Lexical Specificity:} Cyberbullying classes, typically sourced from shorter Twitter posts, often contain explicit, high-signal keywords. This results in higher classification accuracy due to the clear and consistent linguistic markers.
    \item \textbf{Narrative Complexity:} Mental health classes, often from longer, narrative-style Reddit posts, express distress in more nuanced and varied ways. The critical signal may be diffused within a larger volume of text, creating a greater challenge for the model. This aligns with our observation that the most common errors occur between semantically related mental health classes like \textit{Anxiety} and \textit{Stress}.
\end{itemize}

These findings demonstrate the model's ability to adapt to diverse text styles and confirm that performance disparities are a function of the task's complexity, not a reliance on spurious correlations.

\subsection{Limitations and Future Work}
\label{subsec:limitations}
Our work provides a robust framework for multi-class mental health screening, but we acknowledge limitations that pave the way for future research.

\subsubsection{Limitations}
\begin{itemize}
    \item \textbf{Data Provenance and Generalizability:} While we verified label quality using a sample of the datasets, our use of aggregated Kaggle datasets limits broad claims of generalizability. The model's performance on datasets from different platforms, time periods, or demographic groups remains untested.
    \item \textbf{Single-Label Task Framing:}
    Our multi-class framework treats each post as having a single, primary label. This simplifies the clinical reality where mental health conditions and cyberbullying can co-occur (e.g., a post expressing suicidal ideation within a narrative about Bipolar Disorder). As shown in our error analysis (Example 1, Appendix), the model sometimes correctly identifies a high-risk secondary theme (Suicide) while misclassifying the primary label (Bipolar Disorder). A multi-label framework is a critical next step to capture this co-morbidity and provide a more clinically nuanced output.
    \item \textbf{Scope of Evaluation:} The current study is limited to English-language text, excluding the rich signals available in multilingual and multimodal (e.g., images, videos) content.
\end{itemize}

\subsubsection{Future Work}
Based on these limitations, this work highlights several key directions for future research in this domain:

\begin{itemize}
    \item \textbf{Curation of a Multi-Label Benchmark:} The most critical next step for the field is the creation of a new, ethically sourced, and expertly annotated benchmark dataset. Such a dataset should feature multi-label annotations to enable the study of co-occurring conditions, reflecting a more clinically realistic scenario.

     \item \textbf{Validating Utility with Human-in-the-Loop Studies:} The promising results of our prototype pave the way for formal Human-Computer Interaction (HCI) studies. Future work should engage trained moderators in a user study to quantitatively validate the real-world impact of our hybrid explainability system on decision-making accuracy, efficiency, and trust in AI-assisted workflows.

    \item \textbf{Advanced Model Architectures:} The availability of a multi-label benchmark would enable the exploration of multi-task learning architectures. These models could feature a shared encoder with separate classification heads to explicitly model the interplay between different mental health and cyberbullying phenomena.

    \item \textbf{Robustness and Multimodal Evaluation:} Future research should prioritize rigorous cross-domain and temporal generalization tests to ensure models are robust in real-world environments. Furthermore, expanding these frameworks to incorporate multilingual and multimodal analysis is essential for building more comprehensive and inclusive screening tools.
\end{itemize}

\bibliographystyle{unsrt} 
\bibliography{ms}

\begin{thebibliography}{99}

\bibitem{WHO}
World Health Organization. Mental Disorders. 2025. URL: \url{https://www.who.int/news-room/fact-sheets/detail/mental-disorders}. (Accessed: April 22, 2025).

\bibitem{zsila2023}
\'A. Zsila and M. E. S. Reyes. Pros \& cons: impacts of social media on mental health. BMC Psychology, 11(1):201, 2023. DOI: 10.1186/s40359-023-01243-x.

\bibitem{naslund2020}
J. Naslund, A. Bondre, J. Torous and K. Aschbrenner. Social Media and Mental Health: Benefits, Risks, and Opportunities for Research and Practice. Journal of Technology in Behavioral Science, 5, 2020. DOI: 10.1007/s41347-020-00134-x.

\bibitem{acm3653303}
S. Poddar, R. Mukherjee, A. Samad, N. Ganguly and S. Ghosh. MuLX-QA: Classifying Multi-Labels and Extracting Rationale Spans in Social Media Posts. ACM Transactions on Information Systems, 42(3), 2024.

\bibitem{wu2025}
M. Wu, J. Chang, Z. Epstein and D. Rand. Beyond Friends: Exploring the Effects of Unknown Users' Social Media Posts on Individuals' Perceptions and Behaviors. HICSS, 2025. DOI: 10.24251/HICSS.2025.281.

\bibitem{mobin}
Md. I. Mobin, A. F. M. S. Akhter, M. F. Mridha, S. M. H. Mahmud and Z. Aung. Social Media as a Mirror: Reflecting Mental Health Through Computational Linguistics. IEEE Access, 12:130143--130164, 2024. DOI: 10.1109/ACCESS.2024.3454292.

\bibitem{kumar2022}
P. Kumar, P. Samanta, S. Dutta, M. Chatterjee and D. Sarkar. Feature Based Depression Detection from Twitter Data Using Machine Learning Techniques. Journal of Scientific Research, 66(02):220--228, 2022. DOI: 10.37398/JSR.2022.660229.

\bibitem{ji2022mentalbert}
Z. Ji, T. Reddy, A. Nagda and H. Singh. MentalBERT: Publicly available pretrained transformer-based model for mental healthcare social media text mining on Reddit. Proceedings of LREC, 2022.

\bibitem{devlin2018bert}
J. Devlin, M.-W. Chang, K. Lee and K. Toutanova. BERT: Pre-training of deep bidirectional transformers for language understanding. Proceedings of NAACL-HLT, 2019.

\bibitem{devlin2019bert}
J. Devlin, M.-W. Chang, K. Lee and K. Toutanova. BERT: Pre-training of deep bidirectional transformers for language understanding. Proceedings of NAACL-HLT, 2019.

\bibitem{Mathew2020}
B. Mathew, P. Saha, S. M. Yimam, C. Biemann, P. Goyal and A. Mukherjee. HateXplain: A Benchmark Dataset for Explainable Hate Speech Detection. arXiv:2012.10289, 2020.

\bibitem{jiang2020deeppsy}
Y. Jiang. Problematic Social Media Usage and Anxiety Among University Students During the COVID-19 Pandemic: The Mediating Role of Psychological Capital and the Moderating Role of Academic Burnout. Frontiers in Psychology, 12:612007, 2021. DOI: 10.3389/fpsyg.2021.612007.

\bibitem{sosnet2020}
J. Wang, K. Fu and C.-T. Lu. SOSNet: A Graph Convolutional Network Approach to Fine-Grained Cyberbullying Detection. In 2020 IEEE International Conference on Big Data (Big Data), pages 1699--1708, 2020. DOI: 10.1109/BigData50022.2020.9378065.

\bibitem{antypas2023}
D. Antypas and J. Camacho-Collados. Robust Hate Speech Detection in Social Media: A Cross-Dataset Empirical Evaluation. In The 7th Workshop on Online Abuse and Harms (WOAH), Toronto, Canada, Jul 2023, pages 231--242. DOI: 10.18653/v1/2023.woah-1.25.

\bibitem{mentaldata}
S. Sarkar. Sentiment Analysis for Mental Health. Kaggle dataset, 2024.

\bibitem{redditmental}
N. Ghoshal. Reddit Mental Health Data. Kaggle dataset, 2025.

\bibitem{suicidalkaggle}
N. Komati. Suicide and Depression Detection. Kaggle dataset, 2021.

\bibitem{tfidf}
S. Robertson. Understanding Inverse Document Frequency: On Theoretical Arguments for IDF. Journal of Documentation, 60(10):503--520, 2004. DOI: 10.1108/00220410410560582.

\bibitem{wei2019eda}
J. Wei and K. Zou. EDA: Easy Data Augmentation Techniques for Boosting Performance on Text Classification Tasks. arXiv:1901.11196, 2019.

\bibitem{ieee10803545}
A. Nweke, M. Khan and Y. Pei. Explainable Multi-Label Classification Framework for Behavioral Health Based on Domain Concepts. IEEE Transactions on Artificial Intelligence, 15:89--105, 2024.

\bibitem{abdullah2024}
M. Abdullah and N. Negied. Detection and Prediction of Future Mental Disorder From Social Media Data Using Machine Learning, Ensemble Learning, and Large Language Models. IEEE Access, 12:120553--120569, 2024. DOI: 10.1109/ACCESS.2024.3406469.

\bibitem{perez2024}
F. de Arriba-P\'erez and S. Garc\'ia-M\'endez. Detecting anxiety and depression in dialogues: a multi-label and explainable approach. arXiv:2412.17651, 2024.

\bibitem{arxiv2403.10750}
Y. Zhang and J. Liu. Depression Detection on Social Media with Large Language Models. arXiv:2403.10750, 2024.

\bibitem{lundberg2017unifiedapproachinterpretingmodel}
S. Lundberg and S.-I. Lee. A Unified Approach to Interpreting Model Predictions. arXiv:1705.07874, 2017.

\bibitem{agarwal2025gpt}
S. Agarwal et al. gpt-oss-120b \& gpt-oss-20b model card. arXiv preprint arXiv:2508.10925, 2025.

\bibitem{warner2024smarterbetterfasterlonger}
B. Warner et al. Smarter, Better, Faster, Longer: A Modern Bidirectional Encoder for Fast, Memory Efficient, and Long Context Finetuning and Inference. arXiv:2412.13663, 2024.

\bibitem{kojima2022large}
T. Kojima, S. Gu, M. Reid, Y. Matsuo and Y. Iwasawa. Large language models are zero-shot reasoners. Advances in Neural Information Processing Systems, 35:22199--22213, 2022.

\bibitem{mcdermott2024closer}
M. McDermott, H. Zhang, L. Hansen, G. Angelotti and J. Gallifant. A closer look at AUROC and AUPRC under class imbalance. Advances in Neural Information Processing Systems, 37:44102--44163, 2024.

\bibitem{loshchilov2019decoupledweightdecayregularization}
I. Loshchilov and F. Hutter. Decoupled Weight Decay Regularization. arXiv:1711.05101, 2019.

\bibitem{kingma2017adammethodstochasticoptimization}
D. P. Kingma and J. Ba. Adam: A Method for Stochastic Optimization. arXiv:1412.6980, 2017.

\bibitem{liu_sentence_2020}
Z. Liu, Y. Lin and M. Sun. Sentence Representation. In Representation Learning for Natural Language Processing, Springer Nature, 2020, pages 59--89. DOI: 10.1007/978-981-15-5573-2\_4.

\end{thebibliography}

\newpage
\section{Appendix}
\appendix

\section{Model Hyperparameters}
\label{sec:appendix_hyperparams}
This section details the hyperparameters used for all feature extraction methods and machine learning models evaluated in this study. For models trained using cross-validation or grid search, the search space is specified.

\begin{table}[h!]
\centering
\caption{Hyperparameters for Feature Extraction and Trained Models}
\label{tab:hyperparameters}
\resizebox{\columnwidth}{!}{%
\begin{tabular}{lll}
\toprule
\textbf{Component} & \textbf{Model / Method} & \textbf{Hyperparameters} \\
\midrule
\textbf{Feature Extraction} & TF-IDF Vectorizer & Max Features: 5000 \\
                           &                      & Stop Words: None (preserved for training) \\
                           &                      & N-gram Range: (1, 2) \\
                           \cmidrule{2-3}
                           & BERT Embeddings      & Model: \textit{bert-base-uncased} \\
                           &                      & Max Sequence Length: 128 \\
\midrule
\textbf{Lexical Based Models} & Logistic Regression (CV) & Solver: 'saga', Setting: 'multinomial' \\
                             &                          & Max Iterations: 1000 \\
                             &                          & Regularization Strengths (C): [0.1, 1, 10] \\
                             &                          & Cross-Validation Folds: 5 \\
                             \cmidrule{2-3}
                             & Support Vector Machine (Grid Search) & Kernel: 'linear' \\
                             &                          & Regularization Strengths (C): [0.01, 0.1, 1, 10] \\
                             &                          & Cross-Validation Folds: 5 \\
\midrule
\textbf{Static BERT Embedding} & Logistic Regression (CV) & Solver: 'saga', Setting: 'multinomial' \\
\textbf{Based Models}     &                     & Max Iterations: 1000, CV Folds: 5 \\
                           \cmidrule{2-3}
                           & Feedforward NN (DNN) & Hidden Layers: [256, 128] \\
                           &                      & Activation Function: ReLU, Dropout: 0.3 \\
                           &                      & Optimizer: Adam, Learning Rate: 0.001 \\
                           &                      & Batch Size: 64, Epochs: 5 \\
                           \cmidrule{2-3}
                           & Recurrent Neural Network (RNN) & Hidden Layer Dimension: 128 \\
                           &                      & Optimizer: Adam, Learning Rate: 0.001 \\
                           &                      & Batch Size: 64, Epochs: 5 \\
\midrule
\textbf{End-to-End} & Finetuned BERT-base & Optimizer: AdamW \\
\textbf{Fine-Tuned Models} & Finetuned MentalBERT & Learning Rate: 2e-5 \\
                           & Finetuned MentalRoBERTa & Batch Size: 16 \\
                           & Finetuned ModernBERT & Training/Validation Split: 90/10 \\
\bottomrule
\end{tabular}%
}
\end{table}

\section{Model Performances Before Oversampling}
\label{sec:appendix_perf_before}

The performance of the models was evaluated before applying any oversampling. 
The overall and per-class metrics are reported below.

\begin{table}[htbp]
\centering
\small
\caption{Overall ML Model Performance on Non-Oversampled Data}
\label{tab:ml_model_performance_no_os}
\begin{tabular}{lccccc}
\toprule
\textbf{Model} & \textbf{Accuracy} & \textbf{Macro F1} & \textbf{Weighted F1} & \textbf{Precision} & \textbf{Recall} \\
\midrule
TF-IDF\_LogReg & 0.82 & 0.67 & 0.84 & 0.61 & 0.82 \\
TF-IDF\_SVM    & 0.81 & 0.64 & 0.83 & 0.58 & 0.81 \\
BERT\_LogReg   & 0.75 & 0.55 & 0.79 & 0.49 & 0.76 \\
BERT\_RNN      & 0.78 & 0.58 & 0.81 & 0.52 & 0.76 \\
BERT\_DNN      & 0.72 & 0.55 & 0.77 & 0.51 & 0.74 \\
\bottomrule
\end{tabular}
\end{table}

\begin{table}[htbp]
\centering
\scriptsize
\caption{Per-Class Metrics for ML Models on Non-Oversampled Data}
\label{tab:ml_metrics_no_os}
\resizebox{\textwidth}{!}{%
\begin{tabular}{lcccccccccc}
\toprule
\textbf{Model} & \textbf{Age CB} & \textbf{Anx.} & \textbf{Bip.} & \textbf{Eth. CB} & \textbf{Gen. CB} & \textbf{Non-Su.} & \textbf{Pers. Dis.} & \textbf{Rel. CB} & \textbf{Str.} & \textbf{Suic.} \\
\midrule
\multicolumn{11}{l}{\textbf{F1-Score}} \\
\midrule
TF-IDF\_LogReg & 0.89 & 0.57 & 0.45 & 0.86 & 0.67 & 0.85 & 0.43 & 0.93 & 0.19 & 0.87 \\
TF-IDF\_SVM    & 0.92 & 0.56 & 0.39 & 0.86 & 0.63 & 0.84 & 0.29 & 0.93 & 0.17 & 0.85 \\
BERT\_LogReg   & 0.73 & 0.43 & 0.26 & 0.72 & 0.52 & 0.78 & 0.13 & 0.89 & 0.17 & 0.85 \\
BERT\_RNN      & 0.72 & 0.46 & 0.24 & 0.81 & 0.54 & 0.81 & 0.29 & 0.86 & 0.22 & 0.87 \\
BERT\_DNN      & 0.77 & 0.47 & 0.24 & 0.79 & 0.47 & 0.80 & 0.19 & 0.91 & 0.12 & 0.79 \\
\midrule
\multicolumn{11}{l}{\textbf{Precision}} \\
\midrule
TF-IDF\_LogReg & 0.84 & 0.43 & 0.33 & 0.80 & 0.54 & 0.89 & 0.31 & 0.89 & 0.11 & 0.93 \\
TF-IDF\_SVM    & 0.88 & 0.41 & 0.27 & 0.78 & 0.49 & 0.88 & 0.18 & 0.91 & 0.10 & 0.93 \\
BERT\_LogReg   & 0.62 & 0.30 & 0.16 & 0.60 & 0.37 & 0.92 & 0.08 & 0.84 & 0.10 & 0.93 \\
BERT\_RNN      & 0.59 & 0.34 & 0.15 & 0.77 & 0.39 & 0.91 & 0.20 & 0.78 & 0.14 & 0.92 \\
BERT\_DNN      & 0.68 & 0.34 & 0.15 & 0.73 & 0.31 & 0.88 & 0.12 & 0.90 & 0.06 & 0.96 \\
\midrule
\multicolumn{11}{l}{\textbf{Recall}} \\
\midrule
TF-IDF\_LogReg & 0.95 & 0.88 & 0.71 & 0.94 & 0.87 & 0.81 & 0.70 & 0.97 & 0.58 & 0.82 \\
TF-IDF\_SVM    & 0.96 & 0.89 & 0.71 & 0.97 & 0.89 & 0.79 & 0.65 & 0.96 & 0.52 & 0.79 \\
BERT\_LogReg   & 0.89 & 0.76 & 0.65 & 0.89 & 0.87 & 0.68 & 0.50 & 0.96 & 0.60 & 0.79 \\
BERT\_RNN      & 0.91 & 0.74 & 0.71 & 0.85 & 0.87 & 0.73 & 0.50 & 0.96 & 0.46 & 0.82 \\
BERT\_DNN      & 0.87 & 0.78 & 0.58 & 0.87 & 0.93 & 0.73 & 0.45 & 0.91 & 0.66 & 0.68 \\
\bottomrule
\end{tabular}}
\\[4pt]
\textit{CB—Cyberbullying; Anx.—Anxiety; Bip.—Bipolar; Eth.—Ethnicity; Gen.—Gender; Non-Su.—Non-Suicidal; Pers. Dis.—Personality Disorder; Rel.—Religion; Str.—Stress; Suic.—Suicide.}
\end{table}

\begin{table}[htbp]
\centering
\small
\caption{Overall Fine-tuned Model Performance on Non-Oversampled Data}
\label{tab:finetuned_model_performance_no_os}
\begin{tabular}{lccccc}
\toprule
\textbf{Model} & \textbf{Accuracy} & \textbf{Macro F1} & \textbf{Weighted F1} & \textbf{Precision} & \textbf{Recall} \\
\midrule
Fine-tuned BERT       & 0.89 & 0.71 & 0.91 & 0.66 & 0.89 \\
Fine-tuned MentalBERT & 0.90 & 0.73 & 0.92 & 0.66 & 0.89 \\
Fine-tuned RoBERTa    & 0.89 & 0.73 & 0.91 & 0.67 & 0.90 \\
Fine-tuned ModernBERT & 0.88 & 0.69 & 0.90 & 0.63 & 0.88 \\
\bottomrule
\end{tabular}
\end{table}

\begin{table}[htbp]
\centering
\scriptsize
\caption{Per-Class Metrics for Fine-tuned Models on Non-Oversampled Data}
\label{tab:finetuned_metrics_no_os}
\resizebox{\textwidth}{!}{%
\begin{tabular}{lcccccccccc}
\toprule
\textbf{Model} & \textbf{Age CB} & \textbf{Anx.} & \textbf{Bip.} & \textbf{Eth. CB} & \textbf{Gen. CB} & \textbf{Non-Su.} & \textbf{Pers. Dis.} & \textbf{Rel. CB} & \textbf{Str.} & \textbf{Suic.} \\
\midrule
\multicolumn{11}{l}{\textbf{F1-Score}} \\
\midrule
Fine-tuned BERT       & 0.94 & 0.58 & 0.52 & 0.91 & 0.84 & 0.90 & 0.16 & 0.96 & 0.38 & 0.95 \\
Fine-tuned MentalBERT & 0.94 & 0.59 & 0.49 & 0.90 & 0.78 & 0.91 & 0.28 & 0.97 & 0.46 & 0.96 \\
Fine-tuned RoBERTa    & 0.97 & 0.59 & 0.52 & 0.82 & 0.79 & 0.89 & 0.17 & 0.94 & 0.62 & 0.96 \\
Fine-tuned ModernBERT & 0.95 & 0.53 & 0.55 & 0.86 & 0.81 & 0.89 & 0.22 & 0.91 & 0.27 & 0.96 \\
\midrule
\multicolumn{11}{l}{\textbf{Precision}} \\
\midrule
Fine-tuned BERT       & 0.91 & 0.42 & 0.38 & 0.87 & 0.77 & 0.97 & 0.09 & 0.95 & 0.26 & 0.98 \\
Fine-tuned MentalBERT & 0.92 & 0.44 & 0.34 & 0.85 & 0.65 & 0.98 & 0.17 & 0.95 & 0.32 & 0.98 \\
Fine-tuned RoBERTa    & 0.98 & 0.44 & 0.37 & 0.71 & 0.69 & 0.98 & 0.09 & 0.91 & 0.54 & 0.97 \\
Fine-tuned ModernBERT & 0.94 & 0.37 & 0.41 & 0.77 & 0.73 & 0.97 & 0.13 & 0.84 & 0.16 & 0.99 \\
\midrule
\multicolumn{11}{l}{\textbf{Recall}} \\
\midrule
Fine-tuned BERT       & 0.96 & 0.93 & 0.82 & 0.95 & 0.93 & 0.84 & 0.90 & 0.97 & 0.72 & 0.92 \\
Fine-tuned MentalBERT & 0.95 & 0.88 & 0.84 & 0.94 & 0.97 & 0.85 & 0.70 & 0.99 & 0.72 & 0.95 \\
Fine-tuned RoBERTa    & 0.95 & 0.93 & 0.87 & 0.97 & 0.92 & 0.81 & 0.90 & 0.98 & 0.72 & 0.95 \\
Fine-tuned ModernBERT & 0.97 & 0.94 & 0.82 & 0.97 & 0.90 & 0.81 & 0.65 & 0.98 & 0.71 & 0.92 \\
\bottomrule
\end{tabular}}
\\[4pt]
\textit{CB—Cyberbullying; Anx.—Anxiety; Bip.—Bipolar; Eth.—Ethnicity; Gen.—Gender; Non-Su.—Non-Suicidal; Pers. Dis.—Personality Disorder; Rel.—Religion; Str.—Stress; Suic.—Suicide.}
\end{table}

\newpage
\section{Explainability Examples}
\label{sec:appendix_explain}
Below are some examples of our hybrid explainability framework in action. These instances demonstrate how the system provides clear rationales for its predictions. 

\subsection{Prototype Interface Walkthrough}
The ``Social Media Screener'' interface shown in each example is designed for a human-in-the-loop workflow. The key components are:
\begin{itemize}
    \item \textbf{Post Content:} The original text of the post. Words identified by SHAP as having the highest impact on the model's prediction are highlighted in red, providing immediate, quantitative evidence.
    \item \textbf{AI Analysis (Screening Aid Only):} Displays the model's predicted label and confidence score. It includes a critical disclaimer that this is not a clinical diagnosis.
    \item \textbf{LLM Explainability Analysis:} A human-readable narrative, synthesized by an LLM from the SHAP values, explaining \textit{why} the model made its decision in plain language.
    \item \textbf{Moderator Action:} A required action panel where the human user must confirm, dismiss, or re-categorize the flag, ensuring no decision is fully automated.
\end{itemize}

\subsection{Example 1: Misclassification Highlighting Model Nuance}
\noindent\textbf{Text:} \textit{``everything warning i legitimately want to kill myself just to spite my father i lived on my own for many years and about years ago i was guilted into ...''} \\
\textbf{Actual Label:} Bipolar Disorder \\
\textbf{Predicted Label:} Suicidal (\texttt{Confidence: 1.000})

\begin{figure}[htbp]
    \centering
    \includegraphics[width=0.95\linewidth]{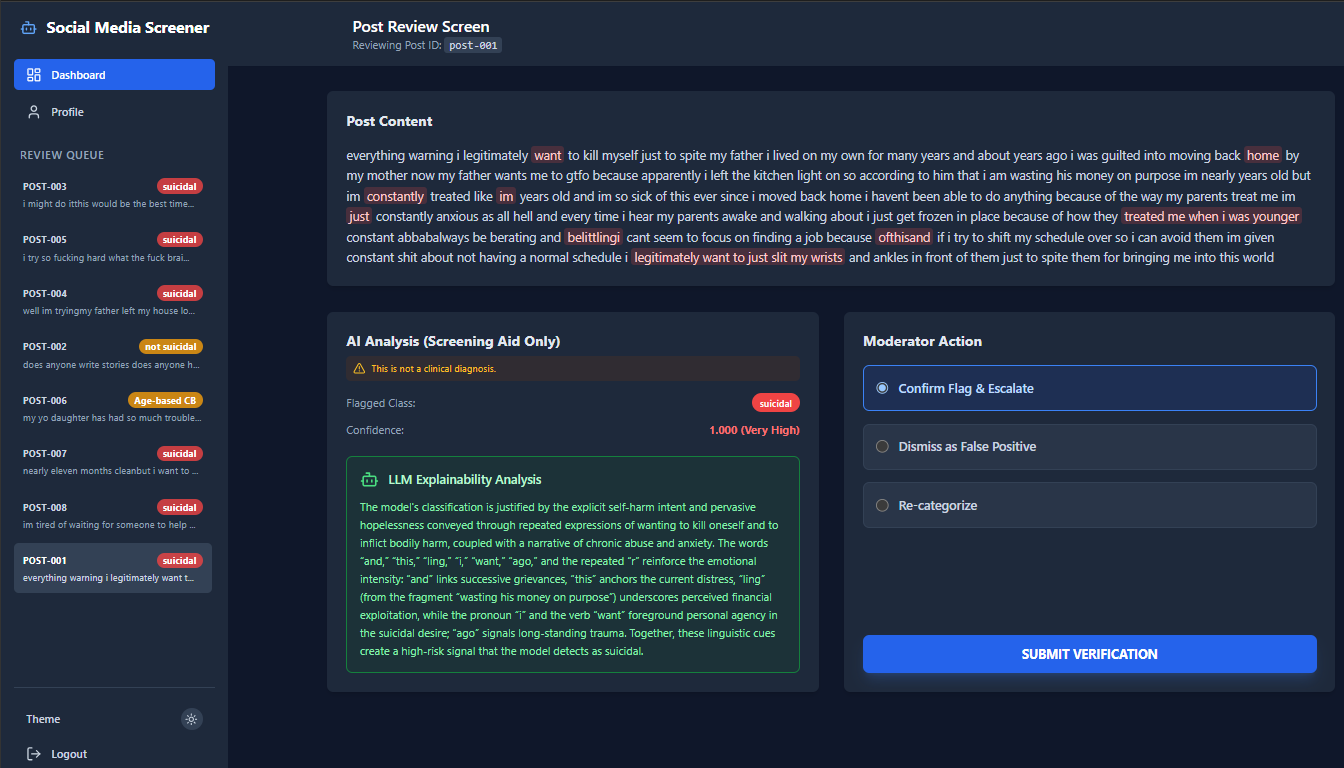}
    \caption{Hybrid explainability output for a text labeled 'Bipolar Disorder' but classified as 'Suicidal' due to strong self-harm intent. This highlights the model's ability to detect co-occurring high-risk language.}
    \label{fig:explain_misclassified}
\end{figure}

\subsection{Example 2: Age-based Cyberbullying}
\noindent\textbf{Text:} \textit{``my yo daughter has had so much trouble at school being left out bullied and the final straw she was jumped by girls on high street of town we live in ...''} \\
\textbf{Actual Label:} Age-based CB \\
\textbf{Predicted Label:} Age-based CB (\texttt{Confidence: 1.000})

\begin{figure}[htbp]
    \centering
    \includegraphics[width=0.95\linewidth]{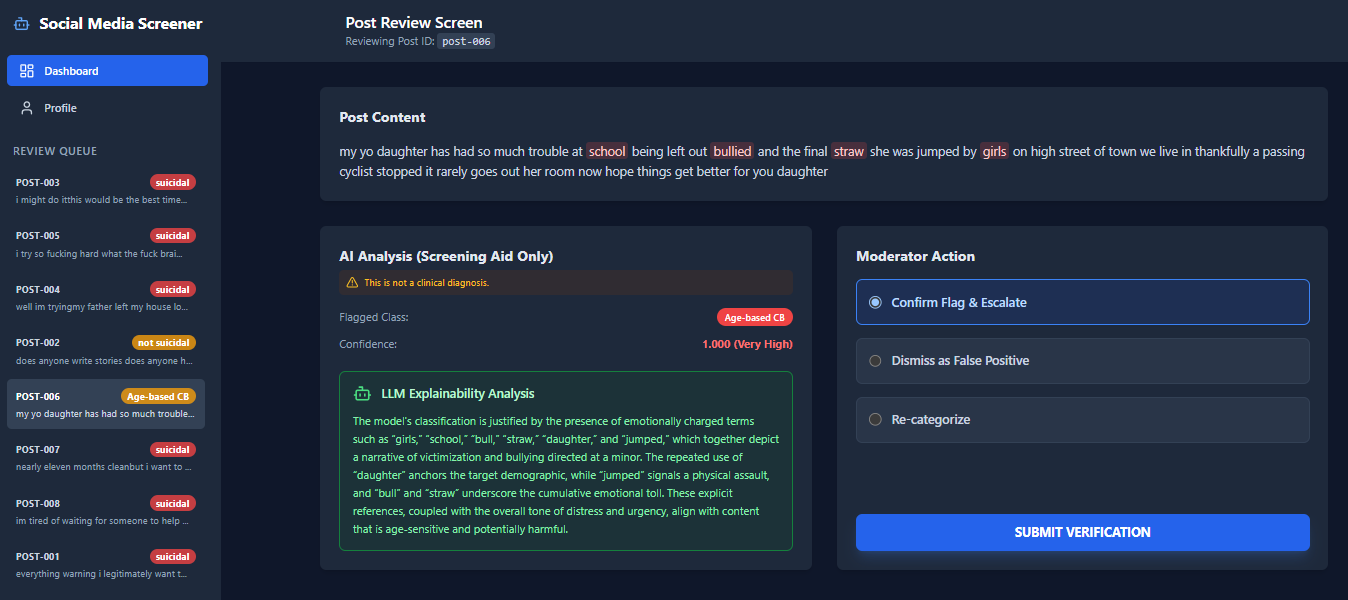}
    \caption{Hybrid explainability output for a text correctly classified as Age-based Cyberbullying. The LLM identifies terms like 'bullied' and 'jumped' as key indicators.}
    \label{fig:explain_age_cb}
\end{figure}

\subsection{Example 3: Suicidal Ideation (Explicit)}
\noindent\textbf{Text:} \textit{``nearly eleven months clean but i want to cut it short by bleeding myself dry in the bathroom fuck i really want to escape whatever the fuck i am whatev...''} \\
\textbf{Actual Label:} Suicidal \\
\textbf{Predicted Label:} Suicidal (\texttt{Confidence: 0.999})

\begin{figure}[htbp]
    \centering
    \includegraphics[width=\linewidth]{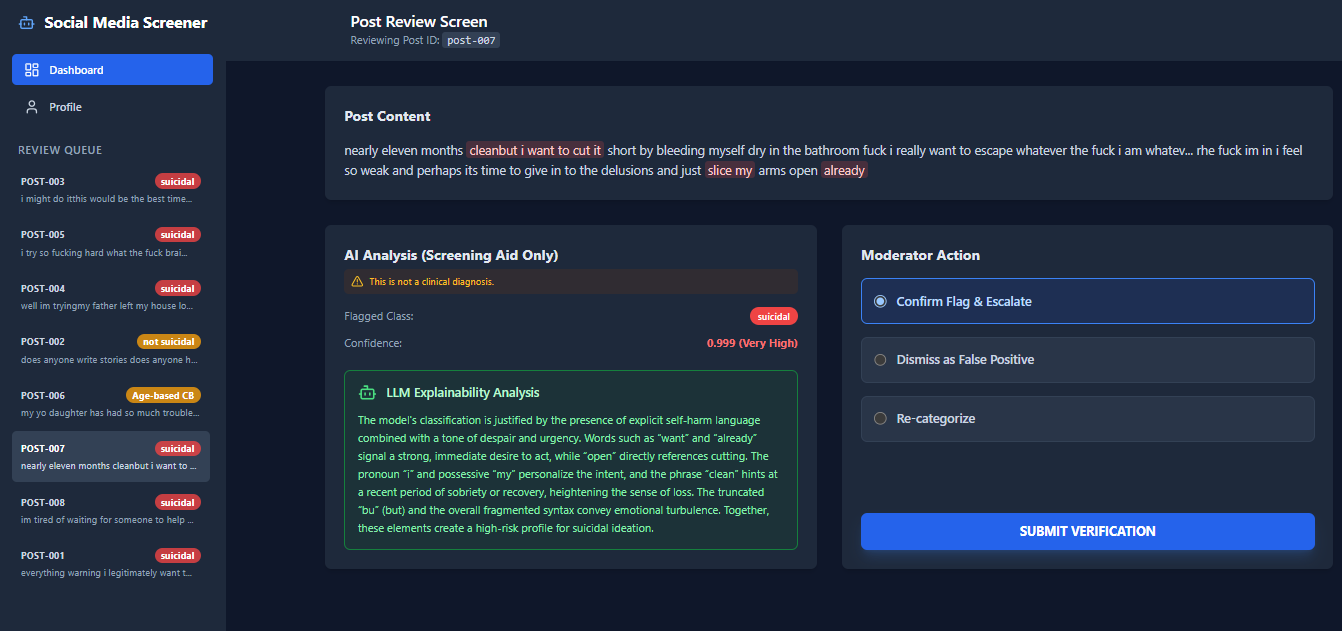} 
    \caption{Hybrid explainability output for a text correctly classified as Suicidal, showing explicit self-harm language ('bleeding myself dry', 'slice my arms').}
    \label{fig:explain_suicide_explicit}
\end{figure}

\subsection{Example 4: Non-Suicidal Content}
\noindent\textbf{Text:} \textit{``does anyone write stories does anyone here write stories with characters is it tough in my experience i feel like i dont know enough about people to write characters who arent just like me...''} \\
\textbf{Actual Label:} not suicidal \\
\textbf{Predicted Label:} not suicidal (\texttt{Confidence: 0.444})

\begin{figure}[htbp]
    \centering
    \includegraphics[width=0.95\linewidth]{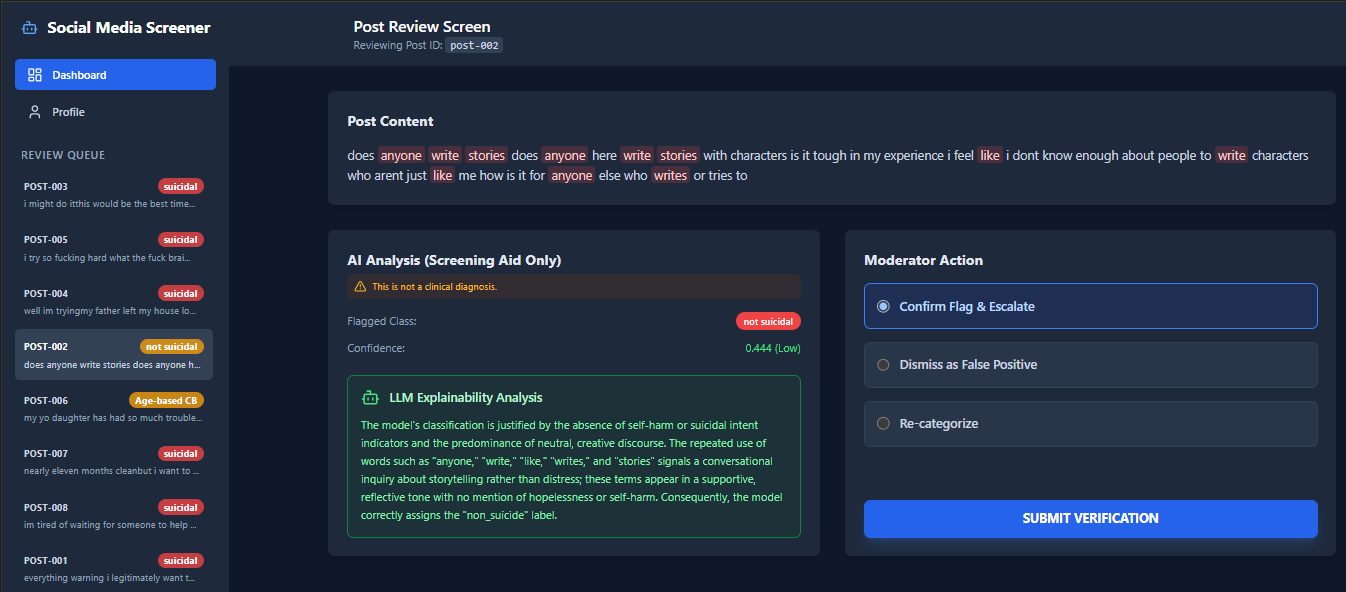} 
    \caption{Hybrid explainability output for a text correctly classified as 'not suicidal'. The LLM notes the absence of self-harm indicators and the focus on creative discussion.}
    \label{fig:explain_non_suicide}
\end{figure}

\subsection{Example 5: Suicidal Ideation (Nuanced)}
\noindent\textbf{Text:} \textit{``i might do this would be the best time considering i lost my closest friends and i have very few people that care about me''} \\
\textbf{Actual Label:} suicidal \\
\textbf{Predicted Label:} suicidal (\texttt{Confidence: 0.996})

\begin{figure}[htbp]
    \centering
    \includegraphics[width=0.95\linewidth]{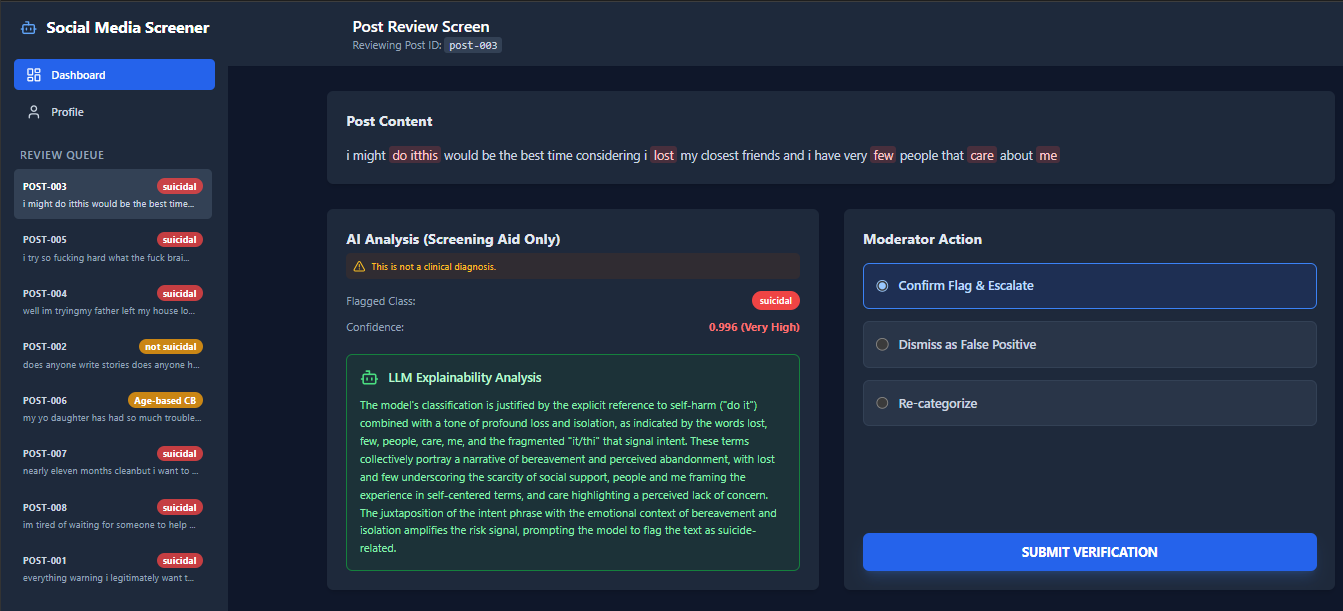} 
    \caption{Hybrid explainability output for a text correctly classified as 'suicidal'. The model identifies nuanced indicators of isolation and hopelessness ('lost', 'few people', 'care about me') as contributing to the risk profile.}
    \label{fig:explain_suicide_nuanced}
\end{figure}
\end{document}